\documentclass[11pt]{article}

\usepackage[]{acl}

\usepackage{times}
\usepackage{latexsym}

\usepackage[T1]{fontenc}
\usepackage[utf8]{inputenc}

\usepackage{microtype}

\usepackage{booktabs}
\usepackage{xcolor}         %
\usepackage{graphicx}
\usepackage{xspace}
\usepackage{todonotes}
\usepackage{multirow}
\usepackage{wrapfig}
\usepackage{caption}
\usepackage{tabularx}
\usepackage{amsmath}
\usepackage{bbm}
\usepackage[ruled,vlined]{algorithm2e}
\usepackage{amsfonts}

\newcommand{\sent}{\texttt{<}$S_1$\texttt{>}}
\newcommand{\firstsent}{\texttt{<}$S_1$\texttt{>}}
\newcommand{\secondsent}{\texttt{<}$S_2$\texttt{>}}
\newcommand{\mask}{\texttt{[MASK]}\xspace}
\newcommand{\entail}{\texttt{entailment}}
\newcommand{\contra}{\texttt{contradiction}}
\newcommand{\nentail}{\texttt{not\_entailment}}
\newcommand{\neutral}{\texttt{neutral}}
\newcommand{\equ}{\texttt{equivalent}}
\newcommand{\nequ}{\texttt{not\_equivalent}}
\newcommand{\posit}{\texttt{positive}}
\newcommand{\negat}{\texttt{negative}}

\newcommand{\ie}{i.e.,\xspace}
\newcommand{\eg}{e.g.,\xspace}
\newcommand{\eat}[1]{}

\newcommand{\baby}{AMuLaP\xspace}

\DeclareMathOperator*{\argmax}{argmax}

\title{Automatic Multi-Label Prompting: \\ Simple and Interpretable Few-Shot Classification} %

\author{Han Wang$^1$\thanks{\ \ Equal contribution.}\ , Canwen Xu$^{2*}$\thanks{\ \ To whom correspondence should be addressed.}\ ,\ Julian McAuley$^2$ \\
$^1$New York University, $^2$University of California, San Diego \\
$^1$\texttt{hwang@nyu.edu}, $^2$\texttt{\{cxu,jmcauley\}@ucsd.edu} \\
}

\begin{document}
\maketitle
\begin{abstract}
Prompt-based learning (i.e., prompting) is an emerging paradigm for exploiting knowledge learned by a pretrained language model. In this paper, we propose Automatic Multi-Label Prompting (AMuLaP), a simple yet effective method to automatically select label mappings for few-shot text classification with prompting. Our method exploits one-to-many label mappings and a statistics-based algorithm to select label mappings given a prompt template. Our experiments demonstrate that AMuLaP achieves 
competitive performance on the GLUE benchmark without human effort or external resources.\footnote{The code is available at \url{https://github.com/HanNight/AMuLaP}.}
\end{abstract}

\section{Introduction}

Since the release of GPT-3~\cite{gpt3}, 
several studies have focused
on exploiting pretrained language models with only a few training examples~\cite{gpt3,gao2020making,autoprompt}. These works demonstrate the 
potential of using natural language prompts to encourage the model to recall 
similar patterns in its training corpus and thus make 
accurate
predictions. This setting of few-shot learning is closer to how humans learn to solve a task, often without many examples as in a traditional deep learning paradigm. The use of prompts can strengthen the explicit connection between input and output, helping the model exploit the knowledge learned from pretraining in a better way. Furthermore, recent works~\citep{schick2021eacl,schick2021naacl,gao2020making} show that prompts can also help the model generalize better in fine-tuning.

Prompt-based learning (\ie prompting) aims to use a template to convert the original input into a prompt-based input with some unfilled masked tokens, and then use the pretrained language model to fill these masked tokens, and finally the tokens filled into these slots are mapped to the corresponding labels as the final output.
In prompting, the design of prompts often plays an important role. Many attempts have been made in this emerging direction of \textit{prompt engineering}~\citep{autoprompt,gao2020making}. Meanwhile, finding a good mapping from the original task labels to tokens (\ie \textit{label engineering}) is also critical to 
few-shot performance, as found in \citet{schick2020coling,gao2020making}. However, manually assigning the label mapping requires human expertise with trial and error. One may argue that the same effort can be used to label more supervised data for a conventional deep learning pipeline. Thus, an efficient automatic label mapping method is 
desirable.

In this paper, we aim to design a method that can automatically find a good label mapping to
save
human effort from label engineering. We propose \textbf{A}utomatic \textbf{Mu}lti-\textbf{La}bel \textbf{P}rompting (\baby), a simple yet effective method to tackle the label selection problem for few-shot classification. \baby is a parameter-free statistical technique that can identify the label patterns from a few-shot training set given a prompt template. \baby exploits multiple labels to suppress the noise and inherently extend the training set for prompt-based fine-tuning. Compared with a hand-crafted label mapping and previous works on automatic label mapping~\citep{schick2020coling,gao2020making}, \baby achieves 
competitive performance despite being simpler and does not require access to the weights of the backbone model, 
or finetune an external pretrained language model for searching label mapping.
We conduct extensive experiments and demonstrate the effectiveness of our method under multiple settings.
Moreover, we attempt to scale \baby with different sizes of the training set and find \baby to work surprisingly well even with one or two shots. We further analyze why does \baby work and discuss the pros and cons of prompting as a new paradigm.

\begin{table*}[t]
\begin{center}
\centering
\small
\begin{tabular}{lllll}
\toprule
\textbf{Task} & \textbf{Template} & \textbf{Class} & \textbf{Manual~\shortcite{gao2020making}} & \textbf{Labels selected by \baby} \\
\midrule
\multirow{3}{*}{MNLI}  & \multirow{3}{*}{{\firstsent} ? {\mask} , {\secondsent}} & \entail & Yes & Yes, Indeed, Also, Currently \\
 & & \neutral & Maybe  & Historically, Suddenly, Apparently, And \\
 & & \contra & No & No, However, Instead, Unfortunately\\
\midrule
\multirow{2}{*}{SST-2} & \multirow{2}{*}{{\sent} It was {\mask} .} & \posit & great & great, perfect, fun, brilliant \\
& & \negat & terrible & terrible, awful, disappointing, not \\
\midrule
\multirow{2}{*}{QNLI}  & \multirow{2}{*}{{\firstsent} ? {\mask} , {\secondsent}} & \entail & Yes & Yes, Historically, Overall, Indeed \\
 & & \nentail & No & Well, First, However, Unfortunately\\
\midrule
\multirow{2}{*}{RTE}   & \multirow{2}{*}{{\firstsent} ? {\mask} , {\secondsent}} & \entail & Yes & Yes, Today, Specifically, Additionally \\
& & \nentail & No & However, Ironically, Also, Indeed \\
\midrule
\multirow{2}{*}{MRPC}  & \multirow{2}{*}{{\firstsent} {\mask} , {\secondsent}} & \equ & Yes & \texttt{</s>}, Currently, Additionally, Today \\
& & \nequ & No & However, Meanwhile, Overall, Finally \\
\midrule
\multirow{2}{*}{QQP}   & \multirow{2}{*}{{\firstsent} {\mask} , {\secondsent}} & \equ & Yes & Or, So, Specifically, Actually\\
& & \nequ & No & Also, And, Finally, Well \\
\midrule
\multirow{2}{*}{CoLA}  & \multirow{2}{*}{{\sent} This is {\mask} .} & \texttt{grammatical} & correct & why, true, her, amazing \\
& & \texttt{not\_grammatical} & incorrect & it, ridiculous, interesting, sad\\
\bottomrule
\end{tabular}
\end{center}
\caption{The manual and automatically selected labels by \baby. The templates used for prompting are from \citet{gao2020making}.\label{tab:templates}}

\end{table*}

\section{Related Work}
\paragraph{Discrete Prompts} The release of GPT-3~\citep{gpt3} has 
led to interest in
\emph{prompting}, a new way 
to leverage
pretrained language models (PLM). \citet{gpt3} proposes an intuitive in-context learning 
paradigm
by concatenating a few input and output examples and feeding them to the language model and let the model autoregressively generate answers for new examples. Recent works~\cite{petroni2019language,davision2019commonsense,jiang2020how} design prompts to probe the factual and common-sense knowledge encoded within a PLM. Recent works~\cite{schick2021eacl,schick2021naacl,gao2020making} demonstrate that even smaller PLMs have similar few-shot learning capacity. \citet{lescao2021naacl} analyzes the effect of prompting and concludes that 
a single
prompt may be worth 100
training examples
in fine-tuning.

Instead of manually designing prompts (\ie prompt engineering), some recent studies also explore automatic prompt generation.  
PETAL~\cite{schick2020coling} augments Pattern Exploiting Training (PET,~\citealp{schick2021eacl,schick2021naacl}) with automatically identified label words; 
\citet{gao2020making} uses re-ranking to find the best label words by fine-tuning a RoBERTa model on the candidates searched by RoBERTa, and using an external generation model for data augmentation of prompt templates;
AutoPrompt~\cite{autoprompt} uses a gradient-based search to determine both prompts and label words. However, these methods require parameter updates with gradient descent, which is infeasible without access to the model weights (\eg GPT-3). 
PET and its variants
also
require a large unlabeled set and need to be fine-tuned multiple times. AutoPrompt uses discretization techniques to approximately map a continuous vector back to tokens in the vocabulary (\ie ``vocablization''). These searched prompts and labels are often uninterpretable by humans.
Different from these prior studies, our proposed \baby is a simple and interpretable method for few-shot prompting that can work well with and without access to 
model weights. Concurrently to our work, \citet{hu2021knowledgeable} propose a method that exploits an external knowledge base to find label mapping.
T0~\citep{sanh2022multitask,promptsource} constructs a dataset of different NLP tasks by manually writing prompt templates and shows that a large language model with multitask training can generalize to unseen tasks.

\paragraph{Continuous Prompts} In parallel with text-based discrete prompts, there is also a line of 
work
focused on tuning only a fraction of parameters of an LM with the help of continuous prompts (\ie soft prompts). \citet{zhong2021naacl} and \citet{qin2021naacl} propose continuous prompts for knowledge probing by tuning some trainable vectors in the input sequence while fixing the rest of the input. \citet{li2021prefix} applies a similar method for natural language generation and achieves comparable performance to fine-tuning while updating only 0.1\% 
of model
parameters. \citet{lester2021power} reveals that prompt tuning is more competitive when scaled up and can achieve 
identical performance to conventional fine-tuning when the model is large enough. \citet{guo2021text} introduces Q-Learning to optimize the soft prompt. Notably, different from discrete prompting, these works often use all training data to update model weights. Different from these works, \baby is a discrete prompting method that has better interpretability and works well in the few-shot setting.

\section{Prompting for Few-Shot Classification}
We follow the setup in LM-BFF~\citep{gao2020making} for few-shot text classification. Given a pretrained language model $\mathcal{L}$, a task $\mathcal{D}$ and its defined label space $\mathcal{Y}$, we have $n$ training examples per class for the training set $\mathcal{D}_{\mathit{train}}$. As pointed out in \citet{perez2021true}, using the \textit{full} development set may be misleading to claim a few-shot setting. Thus, we use a \textit{few-shot} development set with the same size as the training set (\ie $|\mathcal{D}_{\mathit{train}}| = |\mathcal{D}_{\mathit{dev}}|$), to be consistent with \citet{gao2020making} and constitute a ``true few-shot'' setting~\citep{perez2021true}.

For an input example $x$ (a single sentence or a sentence pair), we first use a task-specific template $\mathcal{T}$ to convert it to $x'$, a token sequence with a \mask token. We then map the original label space to a set of selected words from the vocabulary, denoted as $\mathcal{M}: \mathcal{Y}\rightarrow\mathcal{V}'$. Some examples of $\mathcal{T}$ and $\mathcal{M}$ are shown in Table~\ref{tab:templates}. Note that since we focus on automatically finding the label mapping $\mathcal{M}$, we use the manual templates $\mathcal{T}$ from \citet{gao2020making} throughout this paper. Since $\mathcal{L}$ is 
trained to complete the \mask token in an input sequence, we can directly make zero-shot prediction of the probability of class $y \in \mathcal{Y}$ by the masked language modeling: 
\begin{equation}
    p \left( y|x \right) = p \left( \mask = \mathcal{M} \left( y \right) \mid x' \right).
\end{equation}
Alternately,
one can further fine-tune $\mathcal{L}$ with supervised pairs $\left\{ x', \mathcal{M} \left( y \right) \right\}$ to achieve even better performance.

\begin{figure*}
    \centering
    \includegraphics[width=\textwidth]{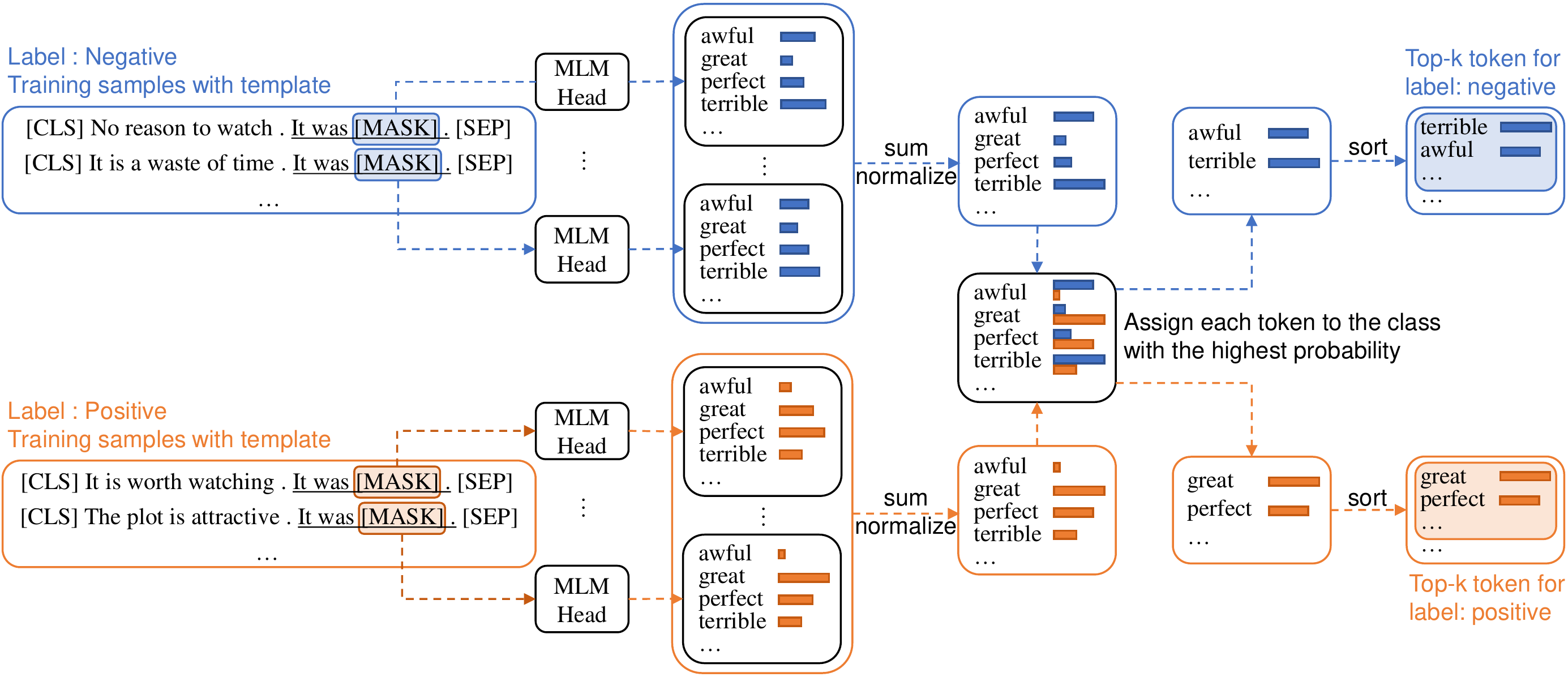}
    \caption{The illustration of implementing \baby on a binary sentiment classification task (SST-2). Each training sample with the task-specific template (the underlined text) is fed into a pretrained language model $\mathcal{L}$ to get its own probability distribution over the vocabulary $\mathcal{V}$. All the obtained probability distributions are summed by class and normalized to get the probability distribution of each class. Then each token in $\mathcal{V}$ is assigned to the class with the highest probability (\eg the token \textit{terrible} is assigned to the class \textit{negative}, the token \textit{great} is assigned to the class \textit{positive}). Finally, for each class, we choose the top-$k$ tokens as label words.}
    \label{fig:amulap}
\end{figure*}

\section{Automatic Multi-Label Prompting}
\subsection{Exploiting Multiple Labels}
Selecting one label word can be insufficient for some complicated tasks, as 
mentioned
in \citet{schick2020coling}. 
We also argue that selecting only one label (especially automatically) may bring noise. This can be resolved by introducing multiple label words. \citet{schick2020coling} use multiple label combinations for PET~\cite{schick2021eacl} and ensemble them afterwards. %
We 
instead
use a straightforward sum to consider multiple label words when making predictions. This design has a similar advantage of exploiting multiple labels without training and ensembling multiple models.

Instead of a one-to-one mapping from the original label space $\mathcal{Y}$ to $\mathcal{V}$, we map each $y \in \mathcal{Y}$ to its label word set $\mathcal{S}(y)$ of $k$ words. We denote the mapping function as $\mathcal{M}': \mathcal{Y} \rightarrow \mathcal{V}^k$. 
For class $y \in \mathcal{Y}$, the predicted probability is calculated as:
\begin{equation}
\label{equ:mulap}
    p \left( y | x \right) = \sum_{v \in \mathcal{S}(y)} p \left( \mask = v \mid x' \right)
\end{equation}
Then, we can simply make predictions by selecting the label with the largest likelihood.

Similarly, if we need to fine-tune $\mathcal{L}$ with supervised pairs, instead of optimizing the cross-entropy loss between the gold label and a single token, we optimize the loss between the sum of the output probabilities of $\mathcal{S}(y)$ and the gold label with a cross-entropy loss:
\begin{equation}
    l = -\sum_{x \in \mathcal{D}_\mathit{train}}\sum_{y \in \mathcal{Y}}\left[\mathbbm{1}\left[y=\hat{y}\right]\cdot\log p\left(y | x\right)\right]
\end{equation}
where $\hat{y}$ is the ground truth label for the input $x$ and $p \left( y | x \right)$ is defined in Equation~\ref{equ:mulap}.

\subsection{Automatic Label Selection}
Finding a good label mapping $\mathcal{M}$ is non-trivial, especially when $\mathcal{M}'$ maps 
an original label to a set of label words instead of one. Selecting a good label mapping often requires 
significant human effort, including domain knowledge and trial-and-error. 
Previously, \citet{schick2021eacl,schick2021naacl} both use hand-crafted label mappings while \citet{schick2020coling} explores automatic label mapping searching but it still requires manual pre-filtering and significantly underperforms the manual mapping. 
\citep{gao2020making} exploits a large pretrained masked language model (RoBERTa, \citealp{liu2019roberta}) to construct a pruned set of label words and then determine the final mapping by fine-tuning on all of them and selecting the best one with $\mathcal{D}_\mathit{dev}$.
We introduce a new selection algorithm for label mapping that achieves competitive results compared to previous efforts.

We aim to achieve two goals: \textbf{(1) Selecting the most likely label mapping based on the training set.} For example, in a 
sentiment
classification task, we would like to see
positive words in the label set of the ``positive'' class while
negative words in the label set of the ``negative'' class. A simple solution is to select 
the
$k$ most likely tokens predicted for the \mask token in the training examples of each class $y$. However, in practice, we would find common words in more than one label set. For example, if we simply take the 10 most likely tokens for the SST-2 dataset~\cite{sst}, we would find ``good'' in both positive and negative label sets, although it is ranked second place in the positive set and ninth in the negative set. Thus, we want to make sure that \textbf{(2) Each token only belongs to at most one label set where it has the highest probability.} To ensure this, we have to iterate over the vocabulary and check that for every token. Then, we can truncate the candidate sets of each class and select the $k$ most likely tokens from each set.
The time complexity of this algorithm is $O(k\cdot|\mathcal{V}|\cdot|\mathcal{Y}|)$.

Formally, we select $\mathcal{M}': \mathcal{Y}\rightarrow \mathcal{V}^k$ by the following steps:
\begin{enumerate}
    \item For each $y_i \in \mathcal{Y}$, we iterate through all training samples $x_j \in \mathcal{D}_\mathit{train}$ whose ground truth label $\hat{y}_j=y_i$. We use $\mathcal{L}$ to predict the token probability of the \mask token and take the average of the predicted probabilities of the $n$ examples to be $\mathbf{z}_i$, where $\mathbf{z}_i$ is a vector over the whole vocabulary.
    \item For each $y_i \in \mathcal{Y}$, initialize an empty candidate token set $\Tilde{\mathcal{S}}(y_i)$.
    \item For each $v \in \mathcal{V}$ where $\mathcal{V}$ is the vocabulary of the model $\mathcal{L}$, we retrieve $v$'s probability value $z_i^v$ from $\mathbf{z}_i$ of each class.
    \item We assign $v$ to the most likely token set of the $m$-th class $\Tilde{\mathcal{S}}(y_m)$ where $m=\argmax_i z_i^v$.
    \item For $y_i \in \mathcal{Y}$, we choose the top-$k$ tokens from $\Tilde{\mathcal{S}}(y_i)$ with the largest probability $z_i^v$ and obtain the truncated word set $\mathcal{S}(y_i)$.
\end{enumerate}

The entire workflow is illustrated in Figure~\ref{fig:amulap}.

\begin{table*}[t]
\centering
\resizebox{1.0\textwidth}{!}{
\begin{tabular}{l|cccccccc|c}
\toprule
    & \textbf{MNLI} & \textbf{MNLI-mm}  & \textbf{SST-2} & \textbf{QNLI} &  \textbf{RTE} & \textbf{MRPC} & \textbf{QQP} & \textbf{CoLA} & \textbf{Avg.} \\
    & (acc) & (acc) & (acc) & (acc) & (acc) & (F1) & (F1) & (Matt.)\\
\midrule
\multicolumn{9}{l}{\textit{\textbf{Baselines}}} \\
\midrule
Majority & 32.7 & 33.0 & 50.9 & 49.5 & 52.7 & 81.2  & 0.0 & 0.0 & 37.5  \\
Manual Label 0-shot~\shortcite{gao2020making} & 50.8  &	51.7 &	83.6  &	50.8  &	51.3  & 61.9  &	49.7  &	2.0 & 50.2   \\
Full Fine-tuning & 89.8 & 89.5 & 95.0 & 93.3 & 80.9 & 91.4 & 81.7 & 62.6 & 85.5\\
\midrule
\multicolumn{9}{l}{\textit{\textbf{Setting 1:} $\mathcal{D}_\mathit{train}$ only; No parameter update.}} \\
\midrule
In-context learning~\shortcite{gpt3}  & \textbf{52.0} (0.7) &	\textbf{53.4} (0.6) &	84.8 (1.3) &	\textbf{53.8} (0.4) &	\textbf{60.4} (1.4) &	45.7 (6.0) &	36.1 (5.2) &	-1.5 (2.4) & 48.1 (2.3) \\
\baby (ours) & 50.8 (2.1) &	52.3 (1.8) & \textbf{86.9} (1.6) & 53.1 (2.8) & 58.9 (7.9) & \textbf{56.3} (5.0) & \textbf{60.2} (2.7) & \textbf{2.3} (1.4) & \textbf{52.6} (3.2) \\

\midrule
\multicolumn{9}{l}{\textit{\textbf{Setting 2:} $\mathcal{D}_\mathit{train}$ + $\mathcal{D}_\mathit{dev}$; No parameter update.}} \\
\midrule

PETAL-CE~\shortcite{schick2020coling} & 48.8 (2.6) &	49.7 (2.3) & 75.6 (7.2)	& 49.5 (0.0) & \textbf{63.5} (3.3) & 28.9 (39.6) & 59.2 (0.0) &	1.3 (3.0) & 47.1 (7.3) \\
PETAL-LR~\shortcite{schick2020coling} & 38.6 (2.0) &	38.4 (2.1) & 85.3 (3.3)	& 53.3 (3.6) & 54.7 (6.4) & 28.0 (38.5) & 55.6 (2.8) &	1.5 (3.4) & 44.4 (7.8) \\
Auto-L~\shortcite{gao2020making} & 41.6 (5.4)	& 42.3 (6.2) & 84.3 (3.3) & 57.9 (3.9) & 61.9 (7.5) & \textbf{67.7} (7.9) & 55.5 (5.0) &	1.2 (4.8) & 51.6 (5.5) \\
\baby (ours) & 50.8 (2.1) &	52.2 (1.9) & 87.0 (1.5)	& 53.5 (2.3) & 59.1 (7.4) & 56.7 (5.7) & \textbf{61.5} (1.7)	& 2.6 (1.8) & 52.9 (3.1) \\
Auto-L + AMuLaP (ours) & \textbf{52.9} (3.0) &	\textbf{54.2} (2.7) & \textbf{90.1} (0.4)	& \textbf{57.9} (2.6) & 59.9 (5.2) & 66.0 (3.0) &	59.4 (2.3) &	\textbf{2.7} (5.7) & \textbf{55.4} (3.1) \\
\midrule
\multicolumn{9}{l}{\textit{\textbf{Setting 3:} $\mathcal{D}_\mathit{train}$ + $\mathcal{D}_\mathit{dev}$; Prompt-based fine-tuning.}} \\
\midrule
Fine-tuning  &	45.8 (6.4) &	47.8 (6.8) &	81.4 (3.8) &	60.2 (6.5) &	54.4 (3.9) & {76.6} (2.5) &	60.7 (4.3) &	\textbf{33.9} (14.3) & 57.6 (6.1) \\
Manual Label FT~\shortcite{gao2020making} &	68.3 (2.3) &	70.5 (1.9) &	92.7 (0.9) &	64.5 (4.2) &	69.1 (3.6) & 74.5 (5.3) &	65.5 (5.3) &	9.3 (7.3) & 64.3 (3.9) \\
PETAL-CE FT~\shortcite{schick2020coling} & 57.5 (3.2) & 57.7 (2.6) & 92.6 (1.0) & 50.5 (0.0) & 68.6 (6.5) & 32.1 (42.5) & 66.7 (3.2) & 3.8 (8.4) & 53.7 (8.4) \\
PETAL-LR FT~\shortcite{schick2020coling} & 64.0 (6.5) & 65.9 (6.4) & 92.9 (1.7) & 65.5 (6.8) & 63.3 (7.7) & 77.7 (3.9) & 65.7 (4.2) & 11.9 (7.5) & 63.4 (5.6) \\
Auto-L FT~\shortcite{gao2020making} & 64.8 (4.7) & 67.3 (4.3) & \textbf{93.5} (0.5) & \textbf{69.8} (3.0) & 67.4 (3.9) & 76.2 (4.8) & 66.4 (4.5) & 23.2 (17.1) & 66.1 (5.4) \\
\baby FT (ours) & \textbf{70.6} (2.7) & \textbf{72.5} (2.4) & 93.2 (0.7) & 65.1 (5.9) & 65.9 (6.3) & \textbf{79.3} (4.0) & \textbf{69.1} (2.5) & 18.3 (9.4) & \textbf{66.8} (4.2) \\
Auto-L + AMuLaP FT (ours) & 68.5 (2.2) & 71.1 (2.3) & 93.4 (1.0) & 69.6 (1.1) & \textbf{69.4} (4.0) & 75.5 (5.6) & 66.4 (3.0) & 14.2 (14.0) & 66.0 (4.2) \\

\bottomrule
\end{tabular}
}

\caption{
Experimental results under three settings with RoBERTa-large as $\mathcal{L}$. For few-shot settings, $n$ is set to 16 per class. We report the average of 5 runs along with their standard deviation in the parentheses.
}
\label{tab:main_results}
\end{table*}

\section{Experiments}
\subsection{Experimental Setting}
\label{sec:settings}
\paragraph{Datasets} We evaluate 
seven classification tasks of the GLUE benchmark~\cite{glue}. Specifically, we test on Microsoft Research Paraphrase Matching (MRPC)~\cite{mrpc}, Quora Question Pairs (QQP) for Paraphrase Similarity Matching; Stanford Sentiment Treebank (SST-2)~\cite{sst} for Sentiment Classification; Multi-Genre Natural Language Inference Matched (MNLI-m), Multi-Genre Natural Language Inference Mismatched (MNLI-mm)~\cite{mnli}, Question Natural Language Inference (QNLI)~\cite{qnli} and Recognizing Textual Entailment (RTE)~\cite{glue} for the Natural Language Inference (NLI) task; The Corpus of Linguistic Acceptability (CoLA)~\cite{cola} for Linguistic Acceptability. We use the manual templates in \citet{gao2020making}, as listed in Table~\ref{tab:templates}. The metrics for each dataset are indicated 
in Table~\ref{tab:main_results}.

\paragraph{Baselines} We compare our method to various baselines: 
\begin{itemize}
    \item \textbf{Majority}: always predict the majority class in the test set.
    \item \textbf{GPT-3-style in-context learning}~\citep{gpt3}: present a few examples to the language model and make it directly predict the next token as the prediction.
    \item \textbf{Manual prompts}: we use the human-designed prompts in \citet{gao2020making}.
    \item \textbf{PETAL-CE}~\citep{schick2020coling}: the variant of PETAL using the cross-entropy metric.
    \item \textbf{PETAL-LR}~\citep{schick2020coling}: the variant of PETAL using the likelihood ratio metric.
    \item \textbf{Auto-L}~\citep{gao2020making}: the automatic label searching method with an external pretrained language model, RoBERTa-large~\citep{liu2019roberta}. The detailed description can be found in Appendix~\ref{sec:appendix}. Note that the results of this baseline is different from those reported in Table 3 of \citet{gao2020making} since they search for both templates and label mapping whereas we fix the templates and search for the label mapping alone, for the sake of fair comparison. We use the officially released code and same hyperparameters for this baseline.
\end{itemize}

\begin{table*}[t]
\begin{center}
\centering
\small
\resizebox{1.0\textwidth}{!}{
\begin{tabular}{cll}
\toprule
\textbf{Class} & \textbf{PETAL-CE}~\cite{schick2020coling}  & \textbf{PETAL-LR}~\cite{schick2020coling} \\
\midrule
\multirow{2}{*}{\posit} & \underline{amazing}, \underline{great}, \underline{brilliant}, \underline{perfect}, \underline{fun}, & \underline{superb}, fearless, \underline{acclaimed}, addictive, visionary, \\
& \underline{wonderful}, \underline{beautiful}, \underline{fantastic}, \underline{awesome}, not & immersive, \underline{irresistible}, timely, unforgettable, \underline{gripping}  \\
\midrule
\multirow{2}{*}{\negat} &  not, \underline{awful}, fun, \underline{funny}, \underline{terrible},  & \underline{annoying}, \underline{insulting}, \underline{meaningless}, \underline{lame}, \underline{shitty}, \\
& great, amazing, hilarious, awesome, good & \underline{humiliating}, childish, \underline{stupid}, \underline{embarrassing}, \underline{irritating} \\
\midrule
\midrule
\textbf{Class} & \textbf{Auto-L}~\cite{gao2020making} & \textbf{\baby} (ours) \\
\midrule
\multirow{2}{*}{\posit} & \underline{exquisite}, \underline{perfection}, \underline{effective}, \underline{fabulous}, intense & \underline{great}, \underline{perfect}, \underline{fun}, \underline{brilliant}, \underline{amazing},  \\
& \underline{inspiring}, \underline{spectacular}, \underline{sublime}, astounding, \underline{thrilling} & \underline{good}, \underline{wonderful}, \underline{beautiful}, \underline{excellent}, \underline{fantastic} \\
\midrule
\multirow{2}{*}{\negat} & \underline{embarrassing}, \underline{boring}, \underline{frustrating}, \underline{ridiculous}, \underline{awkward} & \underline{terrible}, \underline{awful}, \underline{disappointing}, not, \underline{horrible}, \\
& \underline{silly}, nothing, \underline{disgusting}, \underline{ugly}, confusing & obvious, \underline{funny}, inevitable, \underline{bad}, \underline{boring} \\
\bottomrule
\end{tabular}
}
\end{center}
\caption{Most likely label mapping for the SST-2 dataset obtained by PETAL~\citep{schick2020coling}, Auto-L~\citep{gao2020making} and our \baby. Suitable labels annotated by the human annotator are \underline{underlined}.\label{tab:case}}
\end{table*}

\paragraph{Task Setup}
We closely follow the setup in \citet{gao2020making}. We sample $n$ training examples and $n$ development examples per class. We set $k=16$ throughout all experiments. We use RoBERTa-large~\citep{liu2019roberta} as the backbone LM $\mathcal{L}$. For each reported result, we measure average performance across 5 different randomly sampled $\mathcal{D}_\mathit{train}$ and $\mathcal{D}_\mathit{dev}$ splits. Following \citet{gao2020making}, the original development split of each dataset is used as the test set in our experiments. We also report the standard deviation for each result.
To fairly compare with different baselines, we consider the following three settings:
\begin{itemize}
    \item \textbf{Setting 1}: We only use $\mathcal{D}_\mathit{train}$ alone for both label selection and tuning $k$.
    The parameters of $\mathcal{L}$ are not updated. $\mathcal{D}_\mathit{dev}$ is not used. This setting is for fair comparison with \textit{In-context learning}.
    \item \textbf{Setting 2}: We use $\mathcal{D}_\mathit{train}$ for label selection and an additional $\mathcal{D}_\mathit{dev}$ for $k$ tuning. The parameters of $\mathcal{L}$ are not updated. This setting is for fair comparison with Auto-L~\citep{gao2020making} and PETAL~\citep{schick2020coling}.
    \item \textbf{Setting 3}: We use $\mathcal{D}_\mathit{train}$ and $\mathcal{D}_\mathit{dev}$ in the same way as Setting 2 but fine-tune the parameters of the language model $\mathcal{L}$. This setting is for fair comparison with conventional fine-tuning, prompt-based fine-tuning with manual prompts, Auto-L~\citep{gao2020making} and PETAL~\citep{schick2020coling}.
\end{itemize}

\paragraph{Implementation Details} 
We implement \baby based on Hugging Face Transformers~\citep{hf}. When selecting $k$, if there are multiple $k$ with identical performance (which happens occasionally given there are only 16 examples for each class in $\mathcal{D}_\mathit{dev}$), we always choose the largest $k$. For Settings 1 and 2, we search $k$ over $\{1, 2, 4, \ldots, 1024\}$. Note that for settings that do not update the parameters of $\mathcal{L}$, 
search over
$k$ is fast, as we only need to run the model once and cache the distribution of the \mask token. For prompt-based fine-tuning (Setting 3), where we fine-tune the model $\mathcal{L}$, we search $k$ in a smaller space $\{1, 2, 4, 8, 16\}$ due to the increased computational overhead. Following \citep{gao2020making}, we grid search the learning rate from \{1e-5, 2e-5, 5e-5\} and batch size from \{2, 4, 8\}.

\begin{table*}[t]
\centering
\resizebox{1.0\textwidth}{!}{
\begin{tabular}{l|cccccccc|c}
\toprule
    & \textbf{MNLI} & \textbf{MNLI-mm}  & \textbf{SST-2} & \textbf{QNLI} &  \textbf{RTE} & \textbf{MRPC} & \textbf{QQP} & \textbf{CoLA} & \textbf{Avg.} \\
    & (acc) & (acc) & (acc) & (acc) & (acc) & (F1) & (F1) & (Matt.)\\
\midrule

\multicolumn{9}{l}{\textit{\textbf{Setting 2:} $\mathcal{D}_\mathit{train}$ + $\mathcal{D}_\mathit{dev}$; No parameter update.}} \\
\midrule
\baby & \textbf{50.8} (2.1) &	\textbf{52.2} (1.9) & 87.0 (1.5)	& 53.5 (2.3) & \textbf{59.1} (7.4) & \textbf{56.7} (5.7) & \textbf{61.5} (1.7)	& \textbf{2.6} (1.8) & \textbf{52.9} (3.1) \\
\quad w/o dedup. & 45.4 (2.7) &	46.5 (2.5) & \textbf{87.9} (1.0) &	\textbf{53.8} (3.0) & 54.6 (6.0) & 66.7 (12.3) &	57.2 (2.1) &	2.5 (4.2) & 51.8 (4.2) \\
\quad $k=1$ & 46.5 (2.7) & 48.4 (2.6) & 68.8 (12.0) & 51.9 (1.6) & 58.8 (12.7) & 55.0 (4.8) & 59.2 (0.0) & 5.6 (2.1) & 49.3 (4.8) \\
\midrule
\multicolumn{9}{l}{\textit{\textbf{Setting 3:} $\mathcal{D}_\mathit{train}$ + $\mathcal{D}_\mathit{dev}$; Prompt-based fine-tuning.}} \\
\midrule
\baby FT & \textbf{70.6} (2.7) & \textbf{72.5} (2.4) & \textbf{93.2} (0.7) & 65.1 (5.9) & \textbf{65.9} (6.3) & 79.3 (4.0) & \textbf{69.1} (2.5) & 18.3 (9.4) & \textbf{66.8} (4.2) \\
\quad w/o dedup. & 56.9 (5.4) & 58.2 (5.2) & 92.8 (0.9) & 50.6 (0.4) & 57.1 (10.8) & 79.2 (3.6) & 55.0 (26.0) & 5.6 (7.1) & 56.9 (7.4) \\
\quad $k=1$ & 67.7 (4.1) & 69.8 (3.8) & 92.6 (1.0) & \textbf{65.9} (5.2) & 63.1 (8.0) & \textbf{80.2} (3.8) & 66.7 (3.2) & 19.3 (15.5) & 65.7 (5.6) \\
\quad random $\mathcal{M}'$ & 58.8 (6.2) &	61.1 (6.2)	& 92.1 (2.1) &	62.1 (7.1) &	57.0 (11.2) &	74.7 (9.2) &	60.8 (5.8) &	\textbf{31.0} (13.9) & 62.2 (7.7) \\
\quad random $\mathcal{M}'$ ($k=1$) & 52.6 (7.8) &	55.4 (8.3) &	89.0 (4.9) &	65.2 (4.5) &	55.2 (6.2) &	73.4 (10.6) &	60.7 (3.7) &	17.3 (14.7) & 58.6 (7.6) \\

\bottomrule
\end{tabular}
}

\caption{
Experimental results for the 
ablation study. We report the average of 5 runs along with their standard deviation in the parentheses.
}
\label{tab:ablation}
\end{table*}

\subsection{Experimental Results}
We demonstrate experimental results under three settings in Table~\ref{tab:main_results}. Under Setting 1, 
\baby outperforms GPT-3-style in-context learning by 4.5 in terms of the average score and outperforms zero-shot inference with manually designed labels by 2.4. Under Setting 2, compared to variants of PETAL~\citep{schick2020coling}, \baby has an advantage of 5.8 and 8.5 in terms of the average score over CE and LR, respectively. Notably, \baby even outperforms Auto-L by 1.3 without using any external model or data. 
Additionally, we attempt to replace the predicted token distribution of \baby with the validation score of all fine-tuned assignments~\citep{gao2020making}.\footnote{The validation scores of all fine-tuned assignments are obtained on $\mathcal{D}_\mathit{dev}$, as described in \citet{gao2020making}. No external data used. All of these we use are from \url{https://github.com/princeton-nlp/LM-BFF/tree/main/auto_label_mapping}.} 
With the help of many trials in automatic search, 
\baby outperforms Auto-L by a considerable margin of 3.8 in terms of the average score, verifying the versatility of our multi-label mechanism and label selection algorithm. Under Setting 3, \baby FT outperforms all baselines including Auto-L. 
Generally speaking, methods with parameter update (Setting 3) have better performance than those that do not require access to parameters. On all tasks except CoLA, \baby outperforms direct few-shot fine-tuning, suggesting that prompting is a promising method for exploiting large pretrained LMs.

\begin{figure*}
    \centering
    \includegraphics[width=\textwidth]{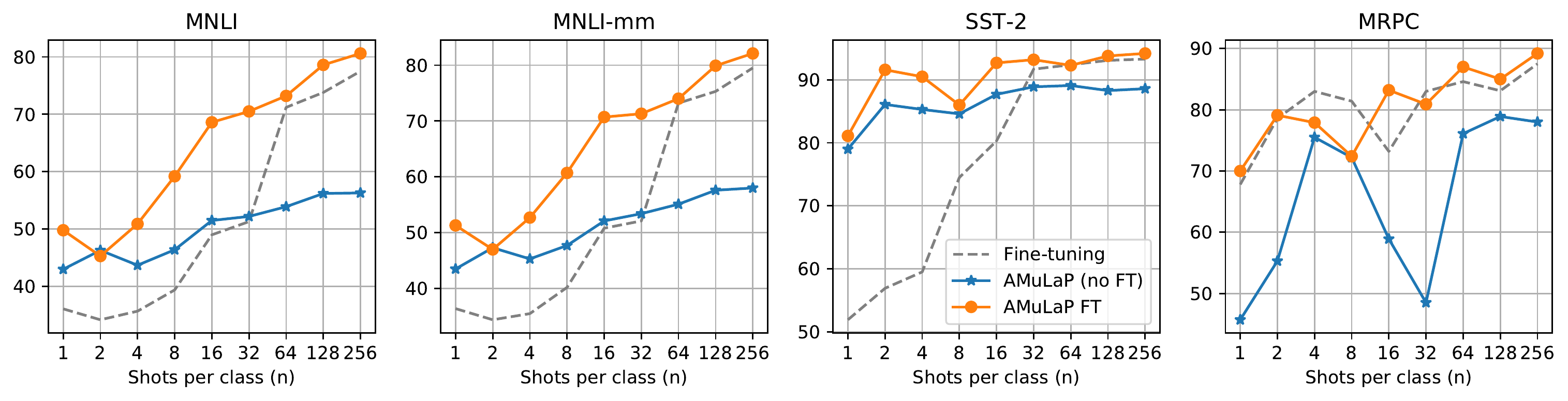}
    \caption{Comparison of \baby, \baby FT and fine-tuning on MNLI, SST and MRPC with different $n$ for the training set and the development set.}
    \label{fig:shots}
\end{figure*}

\section{Analysis}
\subsection{Case Study}
As shown in Table~\ref{tab:case}, we list the 10 most likely label mappings output by PETAL~\citep{schick2020coling}, Auto-L~\citep{gao2020making} and \baby for the SST-2 dataset, respectively. We shuffle the labels from each model and ask a human annotator to annotate whether they are suitable mappings. PETAL-CE suffers from incorrect mappings for ``negative'' while PETAL-LR occasionally outputs vague labels. \baby achieves 
interpretability that is competitive to automatic labels 
obtained by a fine-tuned pretrained language model,
measured by the human agreement ratio. Although \baby outputs three labels that are rated not suitable by the human annotator, it should be noted that all three tokens are ranked low in the candidate set. Thus, introducing top-$k$ truncation can resolve the problem. Additionally, we would like to highlight that \baby mainly collects common words while other methods prefer rare words. This may explain why \baby works well, especially for the non-finetuning settings.

\subsection{Ablation Study}
As shown in Table~\ref{tab:ablation}, we evaluate the effect of each design
choice
on the GLUE benchmark. For both non-finetuning and prompt-based fine-tuning settings, our deduplication algorithm can effectively improve the overall performance by 1.1 and 9.9 in terms of the GLUE average score, respectively. Notably, deduplication is especially important for prompt-based fine-tuning since if the same label maps to two classes,
optimization would be difficult due to the contradiction of supervision signals. Also, our multi-label strategy is shown to be effective
at
improving the average GLUE scores by 3.6 and 1.1 for non-finetuning and fine-tuning settings, respectively. 
Moreover, a random label mapping often leads to lower performance than a label mapping selected based on the training set. An interesting exception is that for CoLA, the random mapping outperforms all label selection methods in Table~\ref{tab:main_results} (both manual and automatic) and is close to the fine-tuning baseline.

\subsection{Scaling 
Few-Shot Learning}
\citet{lescao2021naacl} explore the scaling law of PET~\cite{schick2021eacl} when using more examples for training. Similarly, in this section, we aim to test how \baby scales to different training set sizes $n$.
Figure~\ref{fig:shots} illustrates how standard fine-tuning and our \baby with non-finetuning and fine-tuning compare as $n$ increases. For MNLI and SST-2 task, \baby outperforms standard fine-tuning when we use no more than 16 training examples for non-finetuning and fine-tuning setting. When using more than 16 training examples, \baby under fine-tuning setting still outperforms standard fine-tuning. For an easier task like SST-2, although only 32 training examples are used, the performance of our \baby with non-finetuning and fine-tuning is close to saturation and can be comparable to standard fine-tuning on the entire dataset. For a harder task like MNLI, although the performance of \baby under non-finetuning setting gradually becomes saturated 
as $n$ increases,
\baby under fine-tuning settings continues to improve as $n$ increases and continues to outperform the standard fine-tuning. For MRPC, although the performance of our \baby and standard fine-tuning fluctuate as $n$ increases, in general, \baby with fine-tuning can still achieve comparable performance to standard fine-tuning. In addition, the results demonstrate the effectiveness of \baby especially for extreme few-shot settings. With only one example, \baby achieves decent performance while standard fine-tuning is 
close to random.

\section{Discussion}
\paragraph{Why 
Does \baby Work?}
\citet{schick2020coling} argues that one single label sometimes cannot represent all examples in a class, and thus multiple labels are needed. However, we find this explanation 
insufficient
for understanding the mechanism behind the improved performance with multiple labels. Under a few-shot setting, the limited number of training examples $n$ and complex training procedure of the backbone model $\mathcal{L}$ can often bring 
noise
to both automatic label selection and inference. One example is the meaningless \texttt{</s>} (end-of-sequence marker) label found by \baby, as shown in Table~\ref{tab:templates}. This is due to the format processing in the pretraining of $\mathcal{L}$. Allowing multiple labels can resolve mishaps like this and thus improve the final performance.

Moreover, when selecting multiple labels in fine-tuning, it is equivalent to training on an augmented training set, as multiple labels increase the overall size of the supervision pairs $(x, \hat{y})$. To verify this guess, we test the fine-tuning performance of a random mapping with different labels selected. We find that for random mapping, more labels (\ie a larger $k$) often leads to better performance. This suggests our guess may be correct. However, we do not observe significant improvement when continuing increasing $k$ with labels selected by \baby. As we analyze, increasing $k$ harms the overall quality of selected labels and thus overrides the benefit of a larger $k$. In general, we do not observe a clear law for choosing the best $k$ for \baby. As mentioned before, $k$ can influence both the overall quality of labels (in both ways) and the training procedure (for fine-tuning). Thus, for the optimal performance, we find it essential to search $k$ with a development set.

\paragraph{Limitations and Future Directions} In this paper, we only focus on the selection of the label mapping with a fixed prompt template. There is more to explore when considering the prompt template at the same time. Similar to our paper, previous works~\citep{schick2020coling,gao2020making} separately search for a prompt template $\mathcal{T}$ and the label mapping $\mathcal{M}$. However, these two variables are closely related and greedily search for the best template $\mathcal{T}$ then the best mapping under $\mathcal{T}$ may be suboptimal. 
Jointly searching for $\mathcal{T}$ and $\mathcal{M}$ could be a promising direction for future research.

More broadly, we would like to point out some limitation and contradictions within current few-shot prompting techniques. There is a natural contradiction between performance and access to the model weights. \citet{gpt3} highlights few-shot prompting as a way to mitigate their decision to not release the model weights. However, as shown in our Table~\ref{tab:main_results}, with the same backbone model $\mathcal{L}$, GPT-3-style in-context learning and other methods that do not access the model weights generally underperform those with access to the model weights by a large margin. Also, in-context learning cannot handle more training examples due to the maximum length limit of the model while \baby without fine-tuning gets saturated quickly, as shown in Figure~\ref{fig:shots}.

In addition, complicated prompting techniques are not practically useful for real-world scenarios. For most techniques, the required effort for finding good templates and label mappings, and sometimes training models outweighs the cost of simply labeling more training examples. As shown in Figure~\ref{fig:shots}, 64 examples per class are enough to bring the performance of standard fine-tuning to the same level of prompting. Although recent works on automatic selection of prompts and label mappings are making meaningful contribution to the practicability of few-shot learning, we believe more work should be done to simplify the learning procedure and eliminate human effort while achieving good performance.

\section*{Acknowledgements}
We would like to thank all reviewers for their insightful comments. This project is partly supported by NSF Award \#1750063.

\bibliography{anthology,custom}

\begin{thebibliography}{27}
\expandafter\ifx\csname natexlab\endcsname\relax\def\natexlab#1{#1}\fi

\bibitem[{Bach et~al.(2022)Bach, Sanh, Yong, Webson, Raffel, Nayak, Sharma,
  Kim, Bari, F{\'{e}}vry, Alyafeai, Dey, Santilli, Sun, Ben{-}David, Xu,
  Chhablani, Wang, Fries, AlShaibani, Sharma, Thakker, Almubarak, Tang, Jiang,
  and Rush}]{promptsource}
Stephen~H. Bach, Victor Sanh, Zheng~Xin Yong, Albert Webson, Colin Raffel,
  Nihal~V. Nayak, Abheesht Sharma, Taewoon Kim, M.~Saiful Bari, Thibault
  F{\'{e}}vry, Zaid Alyafeai, Manan Dey, Andrea Santilli, Zhiqing Sun, Srulik
  Ben{-}David, Canwen Xu, Gunjan Chhablani, Han Wang, Jason~Alan Fries,
  Maged~Saeed AlShaibani, Shanya Sharma, Urmish Thakker, Khalid Almubarak,
  Xiangru Tang, Mike~Tian{-}Jian Jiang, and Alexander~M. Rush. 2022.
\newblock Promptsource: An integrated development environment and repository
  for natural language prompts.
\newblock In \emph{{ACL} (Demos)}.

\bibitem[{Brown et~al.(2020)Brown, Mann, Ryder, Subbiah, Kaplan, Dhariwal,
  Neelakantan, Shyam, Sastry, Askell, Agarwal, Herbert{-}Voss, Krueger,
  Henighan, Child, Ramesh, Ziegler, Wu, Winter, Hesse, Chen, Sigler, Litwin,
  Gray, Chess, Clark, Berner, McCandlish, Radford, Sutskever, and
  Amodei}]{gpt3}
Tom~B. Brown, Benjamin Mann, Nick Ryder, Melanie Subbiah, Jared Kaplan,
  Prafulla Dhariwal, Arvind Neelakantan, Pranav Shyam, Girish Sastry, Amanda
  Askell, Sandhini Agarwal, Ariel Herbert{-}Voss, Gretchen Krueger, Tom
  Henighan, Rewon Child, Aditya Ramesh, Daniel~M. Ziegler, Jeffrey Wu, Clemens
  Winter, Christopher Hesse, Mark Chen, Eric Sigler, Mateusz Litwin, Scott
  Gray, Benjamin Chess, Jack Clark, Christopher Berner, Sam McCandlish, Alec
  Radford, Ilya Sutskever, and Dario Amodei. 2020.
\newblock Language models are few-shot learners.
\newblock In \emph{NeurIPS}.

\bibitem[{Davison et~al.(2019)Davison, Feldman, and
  Rush}]{davision2019commonsense}
Joe Davison, Joshua Feldman, and Alexander~M. Rush. 2019.
\newblock Commonsense knowledge mining from pretrained models.
\newblock In \emph{{EMNLP-IJCNLP}}, pages 1173--1178. Association for
  Computational Linguistics.

\bibitem[{Dolan and Brockett(2005)}]{mrpc}
William~B. Dolan and Chris Brockett. 2005.
\newblock Automatically constructing a corpus of sentential paraphrases.
\newblock In \emph{IWP@IJCNLP}.

\bibitem[{Gao et~al.(2021)Gao, Fisch, and Chen}]{gao2020making}
Tianyu Gao, Adam Fisch, and Danqi Chen. 2021.
\newblock Making pre-trained language models better few-shot learners.
\newblock In \emph{{ACL-IJCNLP}}. Association for Computational Linguistics.

\bibitem[{Guo et~al.(2021)Guo, Tan, Liu, Xing, and Hu}]{guo2021text}
Han Guo, Bowen Tan, Zhengzhong Liu, Eric~P Xing, and Zhiting Hu. 2021.
\newblock Text generation with efficient (soft) q-learning.
\newblock \emph{arXiv preprint arXiv:2106.07704}.

\bibitem[{Hu et~al.(2021)Hu, Ding, Wang, Liu, Li, and
  Sun}]{hu2021knowledgeable}
Shengding Hu, Ning Ding, Huadong Wang, Zhiyuan Liu, Juanzi Li, and Maosong Sun.
  2021.
\newblock Knowledgeable prompt-tuning: Incorporating knowledge into prompt
  verbalizer for text classification.
\newblock \emph{arXiv preprint arXiv:2108.02035}.

\bibitem[{Jiang et~al.(2020)Jiang, Xu, Araki, and Neubig}]{jiang2020how}
Zhengbao Jiang, Frank~F. Xu, Jun Araki, and Graham Neubig. 2020.
\newblock How can we know what language models know.
\newblock \emph{Trans. Assoc. Comput. Linguistics}, 8:423--438.

\bibitem[{Le~Scao and Rush(2021)}]{lescao2021naacl}
Teven Le~Scao and Alexander~M. Rush. 2021.
\newblock How many data points is a prompt worth?
\newblock In \emph{{NAACL-HLT}}, pages 2627--2636. Association for
  Computational Linguistics.

\bibitem[{Lester et~al.(2021)Lester, Al-Rfou, and Constant}]{lester2021power}
Brian Lester, Rami Al-Rfou, and Noah Constant. 2021.
\newblock The power of scale for parameter-efficient prompt tuning.
\newblock \emph{arXiv preprint arXiv:2104.08691}.

\bibitem[{Li and Liang(2021)}]{li2021prefix}
Xiang~Lisa Li and Percy Liang. 2021.
\newblock Prefix-tuning: Optimizing continuous prompts for generation.
\newblock \emph{arXiv preprint arXiv:2101.00190}.

\bibitem[{Liu et~al.(2019)Liu, Ott, Goyal, Du, Joshi, Chen, Levy, Lewis,
  Zettlemoyer, and Stoyanov}]{liu2019roberta}
Yinhan Liu, Myle Ott, Naman Goyal, Jingfei Du, Mandar Joshi, Danqi Chen, Omer
  Levy, Mike Lewis, Luke Zettlemoyer, and Veselin Stoyanov. 2019.
\newblock Roberta: {A} robustly optimized {BERT} pretraining approach.
\newblock \emph{arXiv preprint arXiv:1907.11692}.

\bibitem[{Perez et~al.(2021)Perez, Kiela, and Cho}]{perez2021true}
Ethan Perez, Douwe Kiela, and Kyunghyun Cho. 2021.
\newblock True few-shot learning with language models.
\newblock \emph{arXiv preprint arXiv:2105.11447}.

\bibitem[{Petroni et~al.(2019)Petroni, Rockt{\"{a}}schel, Riedel, Lewis,
  Bakhtin, Wu, and Miller}]{petroni2019language}
Fabio Petroni, Tim Rockt{\"{a}}schel, Sebastian Riedel, Patrick S.~H. Lewis,
  Anton Bakhtin, Yuxiang Wu, and Alexander~H. Miller. 2019.
\newblock Language models as knowledge bases?
\newblock In \emph{{EMNLP-IJCNLP}}, pages 2463--2473. Association for
  Computational Linguistics.

\bibitem[{Qin and Eisner(2021)}]{qin2021naacl}
Guanghui Qin and Jason Eisner. 2021.
\newblock Learning how to ask: Querying lms with mixtures of soft prompts.
\newblock In \emph{{NAACL-HLT}}, pages 5203--5212. Association for
  Computational Linguistics.

\bibitem[{Rajpurkar et~al.(2016)Rajpurkar, Zhang, Lopyrev, and Liang}]{qnli}
Pranav Rajpurkar, Jian Zhang, Konstantin Lopyrev, and Percy Liang. 2016.
\newblock Squad: 100, 000+ questions for machine comprehension of text.
\newblock In \emph{{EMNLP}}.

\bibitem[{Sanh et~al.(2022)Sanh, Webson, Raffel, Bach, Sutawika, Alyafeai,
  Chaffin, Stiegler, Raja, Dey, Bari, Xu, Thakker, Sharma, Szczechla, Kim,
  Chhablani, Nayak, Datta, Chang, Jiang, Wang, Manica, Shen, Yong, Pandey,
  Bawden, Wang, Neeraj, Rozen, Sharma, Santilli, Fevry, Fries, Teehan, Scao,
  Biderman, Gao, Wolf, and Rush}]{sanh2022multitask}
Victor Sanh, Albert Webson, Colin Raffel, Stephen Bach, Lintang Sutawika, Zaid
  Alyafeai, Antoine Chaffin, Arnaud Stiegler, Arun Raja, Manan Dey, M~Saiful
  Bari, Canwen Xu, Urmish Thakker, Shanya~Sharma Sharma, Eliza Szczechla,
  Taewoon Kim, Gunjan Chhablani, Nihal Nayak, Debajyoti Datta, Jonathan Chang,
  Mike Tian-Jian Jiang, Han Wang, Matteo Manica, Sheng Shen, Zheng~Xin Yong,
  Harshit Pandey, Rachel Bawden, Thomas Wang, Trishala Neeraj, Jos Rozen,
  Abheesht Sharma, Andrea Santilli, Thibault Fevry, Jason~Alan Fries, Ryan
  Teehan, Teven~Le Scao, Stella Biderman, Leo Gao, Thomas Wolf, and Alexander~M
  Rush. 2022.
\newblock Multitask prompted training enables zero-shot task generalization.
\newblock In \emph{ICLR}.

\bibitem[{Schick et~al.(2020)Schick, Schmid, and
  Sch{\"{u}}tze}]{schick2020coling}
Timo Schick, Helmut Schmid, and Hinrich Sch{\"{u}}tze. 2020.
\newblock Automatically identifying words that can serve as labels for few-shot
  text classification.
\newblock In \emph{{COLING}}, pages 5569--5578. International Committee on
  Computational Linguistics.

\bibitem[{Schick and Sch{\"{u}}tze(2021{\natexlab{a}})}]{schick2021eacl}
Timo Schick and Hinrich Sch{\"{u}}tze. 2021{\natexlab{a}}.
\newblock Exploiting cloze-questions for few-shot text classification and
  natural language inference.
\newblock In \emph{{EACL}}, pages 255--269. Association for Computational
  Linguistics.

\bibitem[{Schick and Sch{\"{u}}tze(2021{\natexlab{b}})}]{schick2021naacl}
Timo Schick and Hinrich Sch{\"{u}}tze. 2021{\natexlab{b}}.
\newblock It's not just size that matters: Small language models are also
  few-shot learners.
\newblock In \emph{{NAACL-HLT}}, pages 2339--2352. Association for
  Computational Linguistics.

\bibitem[{Shin et~al.(2020)Shin, Razeghi, IV, Wallace, and Singh}]{autoprompt}
Taylor Shin, Yasaman Razeghi, Robert L.~Logan IV, Eric Wallace, and Sameer
  Singh. 2020.
\newblock Autoprompt: Eliciting knowledge from language models with
  automatically generated prompts.
\newblock In \emph{{EMNLP}}, pages 4222--4235. Association for Computational
  Linguistics.

\bibitem[{Socher et~al.(2013)Socher, Perelygin, Wu, Chuang, Manning, Ng, and
  Potts}]{sst}
Richard Socher, Alex Perelygin, Jean Wu, Jason Chuang, Christopher~D. Manning,
  Andrew~Y. Ng, and Christopher Potts. 2013.
\newblock Recursive deep models for semantic compositionality over a sentiment
  treebank.
\newblock In \emph{{EMNLP}}.

\bibitem[{Wang et~al.(2019)Wang, Singh, Michael, Hill, Levy, and Bowman}]{glue}
Alex Wang, Amanpreet Singh, Julian Michael, Felix Hill, Omer Levy, and
  Samuel~R. Bowman. 2019.
\newblock {GLUE:} {A} multi-task benchmark and analysis platform for natural
  language understanding.
\newblock In \emph{{ICLR}}.

\bibitem[{Warstadt et~al.(2019)Warstadt, Singh, and Bowman}]{cola}
Alex Warstadt, Amanpreet Singh, and Samuel~R. Bowman. 2019.
\newblock Neural network acceptability judgments.
\newblock \emph{{TACL}}.

\bibitem[{Williams et~al.(2018)Williams, Nangia, and Bowman}]{mnli}
Adina Williams, Nikita Nangia, and Samuel~R. Bowman. 2018.
\newblock A broad-coverage challenge corpus for sentence understanding through
  inference.
\newblock In \emph{{NAACL-HLT}}.

\bibitem[{Wolf et~al.(2020)Wolf, Debut, Sanh, Chaumond, Delangue, Moi, Cistac,
  Rault, Louf, Funtowicz, Davison, Shleifer, von Platen, Ma, Jernite, Plu, Xu,
  Scao, Gugger, Drame, Lhoest, and Rush}]{hf}
Thomas Wolf, Lysandre Debut, Victor Sanh, Julien Chaumond, Clement Delangue,
  Anthony Moi, Pierric Cistac, Tim Rault, R{\'{e}}mi Louf, Morgan Funtowicz,
  Joe Davison, Sam Shleifer, Patrick von Platen, Clara Ma, Yacine Jernite,
  Julien Plu, Canwen Xu, Teven~Le Scao, Sylvain Gugger, Mariama Drame, Quentin
  Lhoest, and Alexander~M. Rush. 2020.
\newblock Transformers: State-of-the-art natural language processing.
\newblock In \emph{{EMNLP} (Demos)}, pages 38--45. Association for
  Computational Linguistics.

\bibitem[{Zhong et~al.(2021)Zhong, Friedman, and Chen}]{zhong2021naacl}
Zexuan Zhong, Dan Friedman, and Danqi Chen. 2021.
\newblock Factual probing is {[MASK]:} learning vs. learning to recall.
\newblock In \emph{{NAACL-HLT}}, pages 5017--5033. Association for
  Computational Linguistics.

\end{thebibliography}
\bibliographystyle{acl_natbib}

\appendix

\section{Automatic Label Selection (Auto-L) in LM-BFF}
\label{sec:appendix}

\citet{gao2020making} proposed a method to automatically construct a label word mapping $\mathcal{M}$ given a fixed template $\mathcal{T}$. They construct a pruned label word set $\mathcal{V}^c \in \mathcal{V}$ of the top $k$ words based on their conditional likehood using the pretrained language model $\mathcal{L}$ for each class $c \in \mathcal{Y}$. They take $\mathcal{V}^c$ as
$$\mathop{\mathrm{Top}\textnormal{-}k}\limits_{v \in \mathcal{V}}\left\{\sum_{x\in\mathcal{D}_{\mathrm{train}}^c}\log p\left(\mask = v \mid \mathcal{T}\left(x\right)\right)\right\}$$
where $\mathcal{D}_{\mathrm{train}}^c \subset \mathcal{D}_{\mathrm{train}}$ denotes the subset of all examples of class $c$. They find the top $n$ assignments over the pruned space that maximize zero-shot accuracy on $\mathcal{D}_{\mathrm{train}}$ to further narrow the search space. Then they fine-tune $n$ assignments and re-rank to find the best label words mapping on $\mathcal{D}_{\mathrm{dev}}$.
\end{document}